\documentclass[conference]{IEEEtran}

\IEEEoverridecommandlockouts                              % This command is only needed if 
\usepackage{algorithm}
\usepackage{algpseudocode}
\usepackage{kotex}
\usepackage{xcolor}
\usepackage{booktabs}
\usepackage{multirow}
\usepackage{xspace}
\usepackage{microtype}

\usepackage{cite}
\usepackage{amsmath}
\usepackage{amssymb}
\usepackage{amsfonts}
\usepackage[final]{graphicx}
\usepackage{textcomp}
\usepackage{etoolbox}
\usepackage{comment}
\usepackage{hyperref}
\usepackage{cleveref}
\usepackage{lipsum}
\usepackage{float}
\usepackage{stfloats}

\usepackage{balance}

\usepackage[T1]{fontenc}

\title{
Understanding Expectations for a Robotic Guide Dog for Visually Impaired People

}

\author{
\IEEEauthorblockN{J. Taery Kim}
\IEEEauthorblockA{\textit{School of Interactive Computing} \\
\textit{Georgia Institute of Technology}\\
Atlanta, USA \\
taerykim@gatech.edu}\\
\IEEEauthorblockN{Bruce N. Walker}
\IEEEauthorblockA{\textit{School of Psychology / Interactive Computing} \\
\textit{Georgia Institute of Technology}\\
Atlanta, USA \\
bruce.walker@psych.gatech.edu}
\and
\IEEEauthorblockN{Morgan Byrd}
\IEEEauthorblockA{\textit{School of Interactive Computing} \\
\textit{Georgia Institute of Technology}\\
Atlanta, USA \\
abyrd45@gatech.edu}\\
\IEEEauthorblockN{Greg Turk}
\IEEEauthorblockA{\textit{School of Interactive Computing} \\
\textit{Georgia Institute of Technology}\\
Atlanta, USA \\
turk@cc.gatech.edu}
\and
\IEEEauthorblockN{Jack L. Crandell}
\IEEEauthorblockA{\textit{School of Interactive Computing} \\
\textit{Georgia Institute of Technology}\\
Atlanta, USA \\
jackcrandell@gatech.edu}\\
\IEEEauthorblockN{Sehoon Ha}
\IEEEauthorblockA{\textit{School of Interactive Computing} \\
\textit{Georgia Institute of Technology}\\
Atlanta, USA \\
sehoonha@gatech.edu}}

\begin{document}

\maketitle
\thispagestyle{empty}
\pagestyle{empty}

%%%%%%%%%%%%%%%%%%%%%%%%%%%%%%%%%%%%%%%%%%%%%%%%%%%%%%%%%%%%%%%%%%%%%%%%%%%%%%%%
\begin{abstract}

% \lipsum[1]
Robotic guide dogs hold significant potential to enhance the autonomy and mobility of blind or visually impaired (BVI) individuals by offering universal assistance over unstructured terrains at affordable costs. However, the design of robotic guide dogs remains underexplored, particularly in systematic aspects such as gait controllers, navigation behaviors, interaction methods, and verbal explanations. Our study addresses this gap by conducting user studies with 18 BVI participants, comprising 15 cane users and three guide dog users. Participants interacted with a quadrupedal robot and provided both quantitative and qualitative feedback. Our study revealed several design implications, such as a preference for a learning-based controller and a rigid handle, gradual turns with asymmetric speeds, semantic communication methods, and explainability. The study also highlighted the importance of customization to support users with diverse backgrounds and preferences, along with practical concerns such as battery life, maintenance, and weather issues. These findings offer valuable insights and design implications for future research and development of robotic guide dogs.

\end{abstract}


%% editing comment

\newcommand{\todo}[1]{\textcolor{blue}{{TODO: #1}}}
\newcommand{\taery}[1]{\textcolor{violet}{{Taery: #1}}}
\newcommand{\sehoon}[1]{\textcolor{red}{{Sehoon: #1}}} 
\newcommand{\wenhao}[1]{\textcolor{blue}{{Wenhao: #1}}} 
\newcommand{\greg}[1]{\textcolor{cyan}{{Greg: #1}}}

\newcommand{\newtext}[1]{#1}
\newcommand{\original}[1]{\textcolor{magenta}{Original: #1}}
\newcommand{\eqnref}[1]{Equation~(\ref{eq:#1})}
\newcommand{\figref}[1]{Figure~\ref{fig:#1}}
\renewcommand{\algref}[1]{Algorithm~\ref{alg:#1}}
\newcommand{\tabref}[1]{Table~\ref{tab:#1}}
\newcommand{\secref}[1]{Section~\ref{sec:#1}}
\newcommand{\mypara}[1]{\noindent\textbf{{#1}.}}

%% ignore text
\long\def\ignorethis#1{}

%% abbreviations
\newcommand{\etal}{{\em{et~al.}\ }}
\newcommand{\eg}{e.g.\ }
\newcommand{\ie}{i.e.\ }

%% reference shortcuts
\newcommand{\figtodo}[1]{\framebox[0.8\columnwidth]{\rule{0pt}{1in}#1}}

%\renewcommand{\eqref}[1]{Equation~(\ref{eq:#1})}

%% frequently used mathematical structures

\newcommand{\pdd}[3]{\ensuremath{\frac{\partial^2{#1}}{\partial{#2}\,\partial{#3}}}}

%% New commands for Sehoon!
\newcommand{\mat}[1]{\ensuremath{\mathbf{#1}}}
\newcommand{\set}[1]{\ensuremath{\mathcal{#1}}}

% math macros
\newcommand{\vc}[1]{\ensuremath{\mathbf{#1}}}
\newcommand{\vEndEff}{\ensuremath{\vc{d}}}
\newcommand{\vRelMove}{\ensuremath{\vc{r}}}
\newcommand{\sSet}{\ensuremath{S}}

\newcommand{\vControl}{\ensuremath{\vc{u}}}
\newcommand{\vPoint}{\ensuremath{\vc{p}}}
\newcommand{\sSpringCoef}{{\ensuremath{k_{s}}}}
\newcommand{\sDamperCoef}{{\ensuremath{k_{d}}}}
\newcommand{\vHandle}{\ensuremath{\vc{h}}}
\newcommand{\vForce}{\ensuremath{\vc{f}}}

\newcommand{\mTransChain}{\ensuremath{\vc{W}}}
\newcommand{\mRotateTrans}{\ensuremath{\vc{R}}}
\newcommand{\sJoint}{\ensuremath{q}}
\newcommand{\vJoint}{\ensuremath{\vc{q}}}
\newcommand{\mJoint}{\ensuremath{\vc{Q}}}
\newcommand{\mMass}{\ensuremath{\vc{M}}}
\newcommand{\sMass}{\ensuremath{{m}}}
\newcommand{\vGravity}{\ensuremath{\vc{g}}}
\newcommand{\vConstr}{\ensuremath{\vc{C}}}
\newcommand{\sConstr}{\ensuremath{C}}
\newcommand{\vCOM}{\ensuremath{\vc{x}}}
\newcommand{\sGeneralForce}[1]{\ensuremath{Q_{#1}}}
\newcommand{\vStateVar}{\ensuremath{\vc{y}}}
\newcommand{\vControlVar}{\ensuremath{\vc{u}}}
\newcommand{\tr}[1]{\ensuremath{\mathrm{tr}\left(#1\right)}}

%%%%%%%%%%%%%%%%%%%%%%%%%%%%%%%%%%%%%%%%%%%%%%%%%%%%%%%%%%%%%%%%%%%
%
% Here are a bunch of macros, mostly for math.
%
%%%%%%%%%%%%%%%%%%%%%%%%%%%%%%%%%%%%%%%%%%%%%%%%%%%%%%%%%%%%%%%%%%%

\renewcommand{\choose}[2]{\ensuremath{\left(\begin{array}{c} #1 \\ #2 \end{array} \right )}}

\newcommand{\gauss}[3]{\ensuremath{\mathcal{N}(#1 | #2 ; #3)}}

\newcommand{\pctab}{\hspace{0.2in}}
\newenvironment{pseudocode} {\begin{center} \begin{minipage}{\textwidth}
                             \normalsize \vspace{-2\baselineskip} \begin{tabbing}
                             \pctab \= \pctab \= \pctab \= \pctab \=
                             \pctab \= \pctab \= \pctab \= \pctab \= \\}
                            {\end{tabbing} \vspace{-2\baselineskip}
                             \end{minipage} \end{center}}
\newenvironment{items}      {\begin{list}{$\bullet$}
                              {\setlength{\partopsep}{\parskip}
                                \setlength{\parsep}{\parskip}
                                \setlength{\topsep}{0pt}
                                \setlength{\itemsep}{0pt}
                                \settowidth{\labelwidth}{$\bullet$}
                                \setlength{\labelsep}{1ex}
                                \setlength{\leftmargin}{\labelwidth}
                                \addtolength{\leftmargin}{\labelsep}
                                }
                              }
                            {\end{list}}
\newcommand{\newfun}[3]{\noindent\vspace{0pt}\fbox{\begin{minipage}{3.3truein}\vspace{#1}~ {#3}~\vspace{12pt}\end{minipage}}\vspace{#2}}

\newcommand{\key}{\textbf}
\newcommand{\fun}{\textsc}

%\def\shortcite{\def\citename##1{}\@internalcite}

% Local Variables:
% TeX-master: "paper"
% End:

% \newcommand{\user}[1]{{
%     \ifthenelse{\equal{#1}{16}} {P#1}{haha}
% }}
\newcommand{\user}[1]{\ifthenelse{\equal{#1}{3} \or \equal{#1}{03} \or \equal{#1}{14} \or \equal{#1}{16}}{P#1-D}{P#1-W}}

\section{Introduction}

% \sehoon{we need a teaser image}

A guide robot has great potential to transform the lives of blind or visually impaired (BVI) individuals by providing autonomous and intelligent mobility guidance in both personal and professional environments. Consequently, researchers have developed various BVI assistance technologies, ranging from smart canes to wheeled mobile robots~\cite{guidecane1997, intellistick2001, spacesene2012, electwalking2015, outdoorfoot2018, tachi1985electrocutaneous, wachaja2017navigating, guerreiro2019cabot, tan2021flying, wang2021navdog, NavBelt1998, HapticBelt2017, SWAN, walkera2021swan}. 
Recent advances have led to the emergence of affordable and capable quadrupedal robots that can traverse rough terrains on relatively low-cost hardware~\cite{7443018, kim2021design, kau2021stanford, a1, kumar2021rma, lee2020, 9714001}. 
Naturally, these quadrupedal robots have attracted the attention of researchers in assistive technology~\cite{xiao2021robotic, chen2023quadruped, defazio2023seeing, morlando2023tethering, hwang2023system, due2023guide, kim2023train, hwang2024towards, wang2022can, hata2024see}, promising ultimate mobility in unstructured environments, including outdoor trails and indoor stairs.

\begin{figure}[t]
    \centering
    \includegraphics[width=0.75\linewidth]{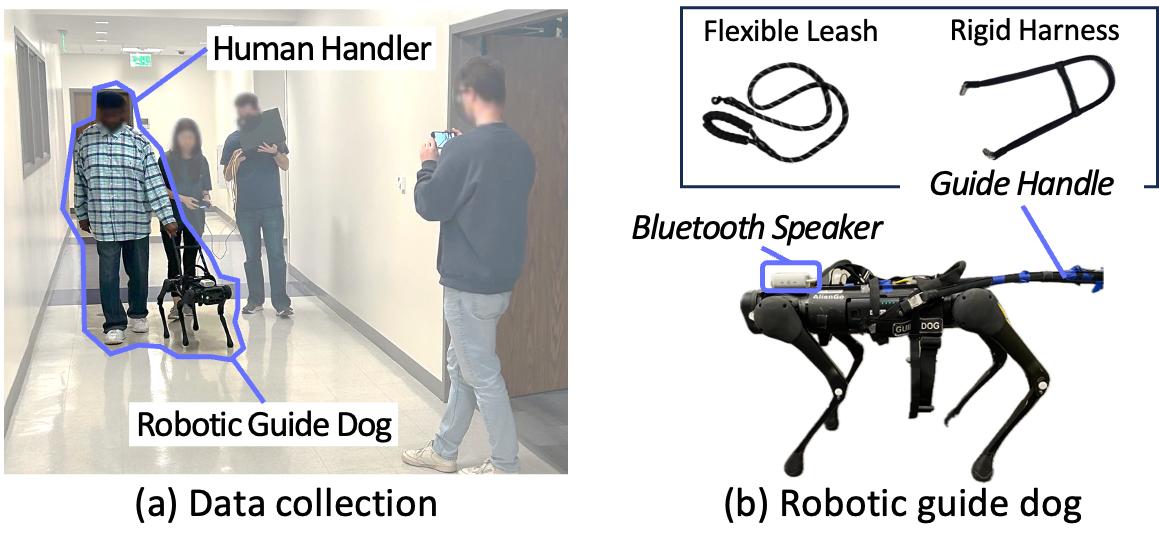}
    \vspace{-1.5em}
    \caption{Overview of the study setup.}
    \vspace{-0.5em}
    \label{fig:teaser}
    \vspace{-1.5em}
\end{figure}

Despite its potential, the design of a robotic guide dog has not yet been fully investigated in the research community, leaving numerous factors underexplored. One possible approach is to simply imitate an animal guide as closely as possible. However, the same design choices may lead to opposite user experiences due to intrinsic differences in sensing and actuation mechanisms. Additionally, a single design choice is associated with multiple attributes, making it difficult for a designer to compare different options. 
For instance, two common gait controllers, model-based optimization and learning-based methods, differ in foot clearance, step lengths, noise levels, and balance recovery strategies, which have their own pros and cons~\cite{9714001, 8594448}. 
Robots can also be equipped with additional features beyond the capabilities of animals~\cite{Setchi2020, Dehkordi2021,
Anjomshoae2019}, such as language-based communication using large language models (LLMs)~\cite{yuan2024towards, padmanabha2024voicepilot, kim2024understanding}, which not only sets them apart from animal guides but also opens up new possibilities for innovation.

We conducted user studies to explore various possible design choices for developing robotic guide dogs. We recruited 18 BVI participants, consisting of 15 cane users and three guide dog users. Each interview session comprised four parts, designed to investigate different aspects of a robotic guide dog: locomotion, navigation movement, physical interaction, and verbal communication. For each part, the participants interacted with a quadrupedal robot and answered questionnaires that gathered both quantitative and qualitative feedback. We analyzed the feedback to understand the expectations and needs of BVI users and summarized possible design implications to inform future researchers interested in exploring guide robots.

Our study identified several design implications, such as a slight inclination toward a learning-based locomotion controller and different preferences for various navigation actions. In addition, the study underscored the substantial need for explainability and personalization to support diverse user preferences. Finally, notable concerns were articulated regarding the practical limitations, such as the battery, maintenance, weather resistance, and emergency issues. 

Our contributions can be summarized as follows: (1) We conducted a user study with 18 BVI users regarding four design aspects of a robotic guide dog, including locomotion, navigation, communication, and explainability.
(2) We identified user preferences for each design factor with statistical analyses.
(3) We concluded the paper with high-level design implications to inform the future design of robotic guide dogs.

\section{Related Work}

\mypara{Participatory Design for Guide Robots}
% -> find anther term for ``participatory design''
% To meet the specific need for assisting BVI individuals, participatory design has been used when developing guide robots.
Designing guide robots requires addressing the practical needs of BVI individuals~\cite{hersh2010robotic}, often approached with participatory design. 
Azenkot et al.~\cite{azenkot2016enabling} used a wizard-of-oz approach to identify primary features of building service robots to assist BVI individuals. % focusing on summoning, choosing assistance modes, and providing building layout information.
Researchers conducted interviews to identify needs in varied settings like shopping malls~\cite{kamikubo2024we}, museums~\cite{kayukawa2023enhancing}, and unfamiliar buildings~\cite{kuribayashi2023pathfinder}.
Researchers have also investigated social aspects~\cite{kayukawa2022users} and autonomy-level preferences for different robot forms~\cite{zhang2023follower, ranganeni2023exploring}
% Kayukawa et al.~\cite{kayukawa2022users} focused on the social aspects of guide robot design, gathering insights from both BVI individuals and bystanders.
% Zhang et al.~\cite{zhang2023follower} and Ranganeni et al.~\cite{ranganeni2023exploring} examined autonomy-level preferences for using car and cane robots. %revealing a need for both full and partial autonomy based on context.
Recent work compared animal and robotic guide dogs to identify key differences and possible opportunities~\cite{due2023guide,hwang2024towards}.
% through a wizard-of-oz approach~\cite{due2023guide} and interaction with animal dog guide handlers and trainers~\cite{hwang2024towards}.
We also engaged with BVI participants to explore multiple design factors of a robotic guide dog.

% \cite{wang2022can} compared legged robot and wheeled robot, resulting in .. 
% However, this work is limited with its default controller, not fully explored the possibility of using legged robot. 
% In our study, we examine different locomotion types in part 1, as well as ..

% Also, previous work tries to predict/model human's movement with respect to robot for planning robot's route. 
% * did *..., 
% These work present good planning, trying to adapt to human speed, etc. focusing on efficiency., but not rarely considering the feelings of human users?
% we test out how users  to the legged robots' movements to , including sidestepping and backstepping

\mypara{Assistive Legged Robots}
% Among the various robot designs proposed to assist BVI navigation, quadrupedal robots have been explored for their similarity to guide dogs. 
% Quadrupedal robots, due to their similarity to dogs, have significant potential to assist BVI individuals in navigation.
% Quadrupedal robots have significant potential for BVI assistance due to their capable mobility over challenging terrains and morphological similarity.
There have been significant advances in quadrupedal control due to the emergence of deep learning and scalable computing frameworks.
Various studies have focused on algorithms for both long-term navigation to a goal point~\cite{xiao2021robotic, chen2023quadruped, cai2024navigating} and short-term navigation for directional cues~\cite{kim2023train, hwang2023system, kim2023transforming, hata2024see}. 
Additionally, force-responsive controls have been developed to enable intuitive interactions with legged robots~\cite{defazio2023seeing, morlando2023tethering}. 
However, many studies prioritize efficiency and rely on a single interaction tool, such as a flexible leash~\cite{xiao2021robotic, chen2023quadruped, defazio2023seeing, morlando2023tethering} or a rigid handle~\cite{kim2023train, hwang2023system, kim2023transforming, hata2024see}, often with little consideration for user preferences.

Quadrupeds show promise for BVI assistance due to their mobility over challenging terrains and morphological similarity to dogs. Yet, their use as robotic guides remains debated.
Wang et al.~\cite{wang2022can} highlighted that noise from legged robots can interfere with users' ability to sense their surroundings.
Due et al.~\cite{due2023guide} questioned legged robots' capabilities compared to animal guides, noting issues from the robots' self-adjustments and limited responsiveness. 
However, these studies were limited by their use of default controllers without fully exploring the potential of legged robots. 
Our work aims to address this gap by examining diverse locomotion and navigation controllers.
% and incorporating user feedback to optimize the design and behavior of quadrupedals for guiding purposes.

%%SH.7.11: Move this to Part 4.
% \mypara{Explainable Agent for Human-Robot Interaction}
% % explainable agent, hci
% Explainability in human-robot interaction refers to the robot’s ability to explain its behavior. It helps the human user to understand the mental model of the robot so that they can have reasonable expectations of the robot’s behavior and limitations and can trust its actions \cite{Setchi2020}, \cite{Dehkordi2021}. Please see the survey paper \cite{Anjomshoae2019} to see a detailed description of current research. Our work seeks to determine the benefits of audio explanations for improving the user experience when using a robot dog guide.

\section{User-study Design}
We used a quadrupedal robot AlienGo from Unitree~\cite{aliengo} as our robot platform (Fig.~\ref{fig:teaser}).
% Robot image?
% See Appendix 1 in AlienGo Software Guide PDF
% Weight = 25kg, size (L,W,H) = (0.55, 0.35, 06) m when standing, (0.55, 0.35, 0.31) m when walking
This robot, weighing 25 kg with dimensions (L, W, H) of 0.55 m, 0.35 m, and 0.6 m when standing, is comparable in size to an average Labrador Retriever, a typical guide dog breed.
The robot was equipped with two types of off-the-shelf handles: a default option of a rigid harness handle designed for guide dogs and a flexible leash for comparison. 
The leash was adjusted to match the harness length and attached to the same point on the robot for a fair comparison.
% Handle is 0.4 m or 0.5 m. I'll need to check it to see which type it is.
% I guess I can measure the leash if that is useful or required.
For the explainability part of our study, a Bluetooth speaker was placed on top of the robot's head.
% (device info needed?)

% \subsection{BVI Participants and IRB Approval}
\begin{table}[t]
    \caption{Demographic information of the BVI participants\\(* hearing impairment; ** owned guide dog before)}
    \scriptsize
    \vspace{-1em}
    \centering
    \setlength\tabcolsep{2.5pt}
    \begin{tabular}{cccccc}
    % \hline
    \toprule
        \textbf{ID} & \textbf{Gender} & \textbf{Age} & \textbf{Impairment Condition} & \textbf{Mobility Aid}\\ \midrule
        \user{1} & M & 67 & Low vision & White Cane\\
        \user{2} & M & 64 & No sight in right, low vision in left & White Cane\\
        \user{3} & M & 56 & Low vision & Guide Dog\\
        \user{4} & F & 64 & Blind & White Cane\\
        \user{5} & F & 51 & No sight in right, low vision in left 
        % (Glaucoma, Diabetic sight)
        & White Cane\\ 
        % \addlinespace
        \user{6} & M & 77 & Blind & White Cane\\
        \user{7} & F & 51 & Blind & White Cane**\\
        \user{8} & F & 68 & {Low vision*} & White Cane\\
        % P08 & 68 & F & \textls[-1]{Low vision, light perception*} & White Cane\\
        \user{9} & M & 67 & {Low vision*} & White Cane\\
        % P09 & 67 & M & \textls[-50]{Low vision, low depth perception*} & White Cane\\
        % P09 & 67 & M & \scalebox{.9}[1.0]{Low vision, low depth perception; hearing loss} & White Cane\\
        \user{10} & M & 51 & Blind & White Cane\\
        % P10 & 51 & M & Blind (Diabetic Retinopothy) & White Cane\\
        % \addlinespace
        \user{11} & F & 37 & Low vision & White Cane\\
        % P11 & 37 & F & Low vision (Uveitis, Retinal detachment) & White Cane\\
        \user{12} & F & 64 & Low vision & White Cane\\
        \user{13} & F & 34 & Low vision & White Cane\\
        \user{14} & M & 69 & Blind & Guide Dog\\
        \user{15} & F & 68 & Low vision & White Cane\\ 
        % \addlinespace
        \user{16} & F & 62 & \shortstack{No sight in left, low vision in right*} & Guide Dog\\
        \user{17} & F & 66 & Low vision* & White Cane\\
        \user{18} & M & 46 & Low vision & White Cane\\
        \bottomrule
    \end{tabular}
    \label{tab:bvi-info}
    \vspace{-2.0em}
\end{table}

We recruited 18 visually impaired participants who use mobility aids daily, as summarized in Table~\ref{tab:bvi-info}.
Our participants consisted of 8 males and 10 females, with ages ranging from 34 to 77.
Three participants used a guide dog as a current mobility aid, and the others used a white cane.
% owned before; had guide dog training;

This study was approved by the university's ethics review panel (IRB).
After obtaining consent from each participant, we underwent the familiarization phase in a lab space.
For those without prior experience with guide dogs, we briefly introduced the concept of guide work.
Participants were then asked to hold the harness handle with their left hand and walk on the right side of the robot, as human handlers typically do with guide dogs~\cite{dietz2003look, alterisio2019you, pietrowiak2024follow}.
The forward walking speed of the robot was adjusted to match each participant's gait during this phase.
Once the participants were comfortable with the guide robot interaction, we proceeded with the study.
% which included four parts: 1. Locomotion, 2. Human Movement, 3. Communication Modality, and 4. Explainability.
% We used the wizard-of-oz method throughout this study to understand ``what the users should do'', ``what the user will do'', and ``what the robot should do''~\cite{green2004applying}.
We used the wizard-of-oz method throughout this study to understand ``what the user will do'', and ``what the robot should do''~\cite{green2004applying}.

% Stat analysis method
For all data, we tested normality using the Shapiro-Wilk test~\cite{shapiro1965analysis} and determined non-parametric tests were appropriate for all parts.
The Wilcoxon signed-rank test was used for paired data in Part 1, 3, and 4, while the Friedman test, Nemenyi post-hoc test, and Spearman's rank test were used in Part 2.

\section{Part 1. Robot Locomotion}\label{sec:p1-locomotion}
We designed experiments to understand the influence of gait controllers. Prior studies~\cite{wang2022can} have suggested that quadrupedal robots might not be suitable for assisting BVI individuals due to the distinctive noise produced during their gaits, which could potentially impede users' perception of other important sounds in the environment. However, those studies primarily relied on company-provided model predictive control (MPC) without considering broader gaits with various styles.
% , such as learning-based locomotion controllers. 
% \sehoon{Taery, add citations that says disadvantages of robot guide dogs}

This section aims to address our key hypothesis that users will have different experiences based on the characteristics of the robot gaits. We focus on two factors: noise level and navigation compliance. The former refers to the volume of noise generated by foot-ground contacts, while the latter describes how a robot sensitively changes its intended trajectory with respect to external disturbances (e.g. obstacles, people). Then, we will discuss user preferences based on their characteristics.
% The first part is understanding how the users perceive the robot's gait characteristics.
% Legged robots tend to have stepping noise on top of fan noise from computing machines
% unlike wheeled robots.
% We use two locomotion policies for this experiment, focusing on two characteristics: noise level and compliance-resistance level.
% \sehoon{Motivate why this is an issue - cite a paper. }
% \sehoon{Briefly introduce two gaits- philosophies and high-level characteristics.}
% \begin{itemize}
%     \item Is the noise level from the robot gait acceptable? How much noise is expected from the robot?
%     \item Does the guide robot should have higher compliance or resistance?
% \end{itemize}

% what do we want to learn from this experiment?
% white cane or mobility aids with wheel, user has more control

\vspace{-0.3em}
\subsection{Procedure}
\vspace{-0.5em}
\mypara{Locomotion policies}
We implemented two gait controllers: (1) model-based and (2) learning-based. For the model-based controller, we employed the default controller provided by the vendor, which is associated with higher noise and compliance levels. On the other hand, the second learning-based policy was trained using deep reinforcement learning \cite{schulman2017proximalpolicyoptimizationalgorithms} in a massively parallelized simulation framework \cite{makoviychuk2021isaac}. The framework was further augmented by an extra sound-reduction reward component aimed at penalizing excessive contact forces. As a result, the policy causes the robot to walk with reduced noise.

Note that our gait controllers are selected to exhibit different characteristics, but they do NOT illustrate the limits of either model-based or learning-based approaches in general. In theory, both can be tuned to produce specific gait styles.

\mypara{Study procedure}
% Initially, users were introduced to the two gaits for familiarity with our experimental setup. 
To familiarize participants with our experimental setup, they were initially introduced to the two gaits, which were referred to as \textit{first} and \textit{second} rather than by name to prevent bias.
After this familiarization phase, the participants were asked to navigate a route twice, with the order of the two gaits counterbalanced across participants to mitigate order effects. The route consisted of an environmental change in indoor settings from a carpeted lab space to a tiled hallway with five turns. After the experiments, we asked the participants to elaborate on the observed differences and rate preference, noise level, and compliance level on a 5-point Likert scale. 
% Upon the request, we allowed users to have additional interactions with the robot in the lab space, such as tugging, pulling, and pushing, on random trajectories.
Upon request, we allowed users to have additional interactions with the robot, such as tugging, pulling, and pushing, during random trajectories in the lab space.

% After the familiarization phase, the participants are asked to follow a route twice with two types of gait.
% The order of gait is randomized for each participant.
% The route consists of an environmental change in indoor settings from a carpeted lab space to a tiled hallway with five turns.
% After the experiments, we asked the participants to elaborate on the differences they noticed and to rate preference, noise level, and compliance level.

% For those who had difficulty answering the compliance level questions, we asked the participants to follow the robot's guide in a lab space while actively interacting with the robot.
% The interactions consisted of tugging, pulling, and pushing while the robot made random movements such as pausing, moving forward, or turning in different directions.

\vspace{-0.5em}
\subsection{Results}
\vspace{-0.5em}
\begin{figure}
    \centering
    \includegraphics[width=0.9\linewidth,trim={0 0 0 2cm}, clip]{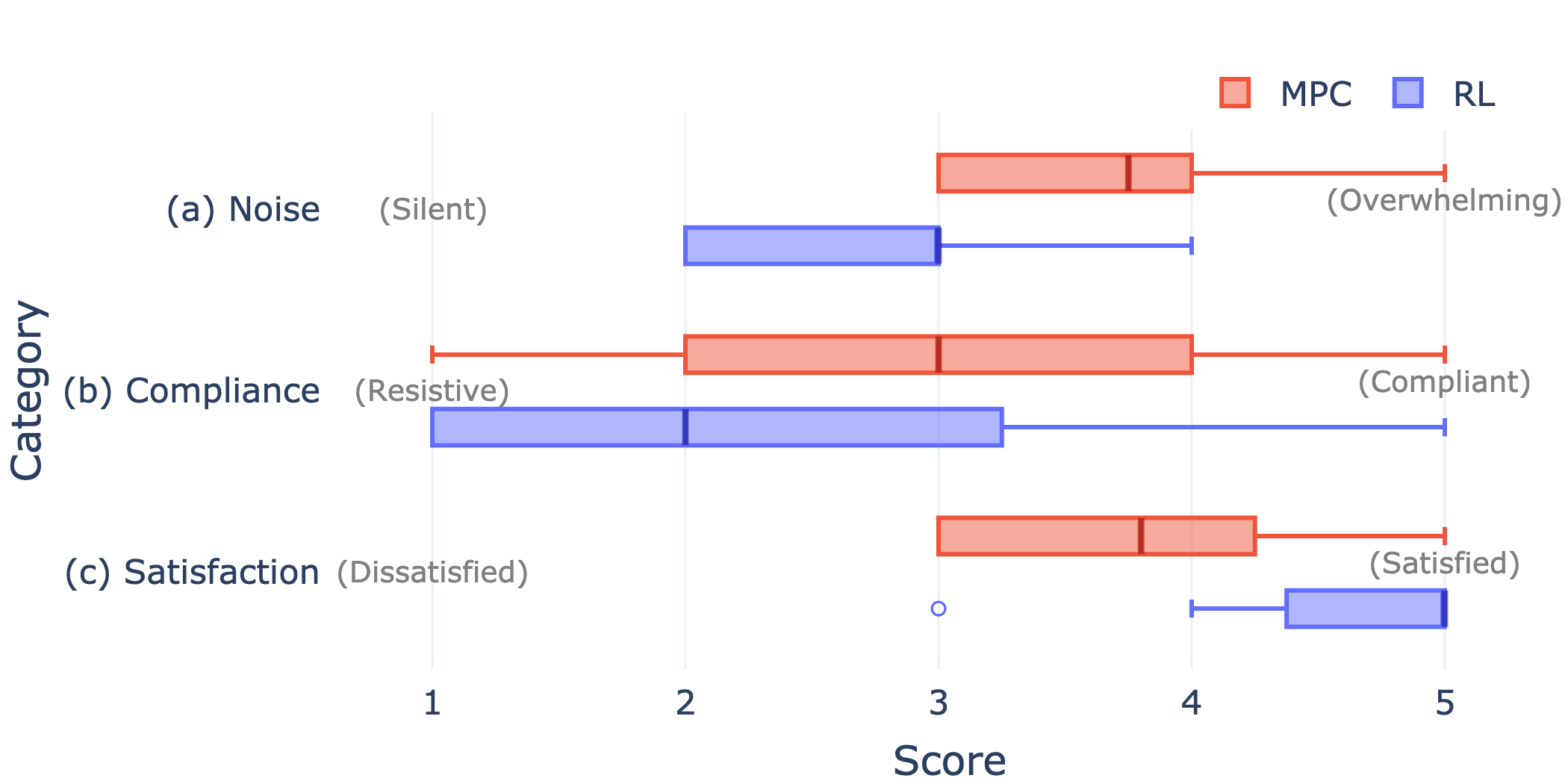}
    \vspace{-1.0em}
    \caption{Comparison of ratings for two types of robot locomotion, Model Predictive Control (MPC, red) and Reinforcement Learning (RL, blue). 
    % The box plots illustrate the distribution scores for noise, resistance, and satisfaction. 
    (a) Noise (silent to overwhelming): MPC demonstrates a generally higher noise level than RL. 
    (b) Compliance (resistive to compliant): RL shows wider variation. 
    (c) Satisfaction (dissatisfied to satisfied): RL shows tighter clustering with a higher median, indicating more consistent satisfaction than MPC.
    % \sehoon{Explicit about gait names. Fix an overlap between texts.}
    % \taery{resistance -> compliance?}\sehoon{yes}
    }
    \label{fig:q1-gait}
    \vspace{-1.75em}    
\end{figure}

\mypara{Overall impression}
With the exception of \user{16}, all noted significant differences between the two locomotion styles with many expressing a clear preference immediately following the trials. An MPC gait is characterized by frequent, harsh, and noisy steps, making it somewhat challenging for users to keep up with its pace. In contrast, an RL gait is perceived as more natural and smooth, facilitating easier tracking for users.
% All the participants, except for P16, found the two locomotions very different after trying out each once.
% Most of the participants specified their preference right after the trials and described how those were different.
% For MPC, the participants mentioned that it was stepping a lot more and not really going anywhere; seemed like a puppy; a bit more difficult to keep up with/flow with; hasty and harsh steps; bumped into them.
% On the contrary, RL was described to be walking it was going somewhere; seemed "seasoned", not a puppy; more like a trot, easier to keep up with; more natural/smoother.
% \sehoon{Yes, the gait seems to be an important factor that affects UX}
% As a common comments for two gaits, both were slow, doesn't seem to turn as tightly, 

\mypara{Reduced noise of RL gaits}
The participants identified that the MPC gait exhibited larger noise scores (average: 3.64) compared to the RL gait (average: 2.81). Fourteen participants rated MPC as noisier, while four participants assessed a similar noise level of 3 (`acceptable') for the two controllers. 

Despite the consistent ratings, participants held slightly varied opinions regarding the desired noise level, which is somewhat against the findings in previous studies~\cite{wang2022can}. The majority expressed a preference for a more quiet gait to avoid attracting attention or disrupting others, particularly in quiet environments such as offices. Notably, participants with additional hearing impairments unanimously favored minimal sound production to better focus on their surroundings, with \user{9} noting an enhanced sense of safety with a quieter robot.
% during the trial with lower noise levels.

Interestingly, some participants favored a robot emitting some sound, as it aids in perceiving its operational status and location, which can be beneficial in noisy or crowded environments like train or subway stations, as well as shopping malls.
\user{7} noted that ``some sound is useful, so you can know where the robot is,'' and \user{13} even suggested that it could give information on ``changing textures or terrains.''
They also highlighted that a proper noise level could reduce their mental effort and the chance of losing track of the robot. However, this group also emphasized the importance of the robot's ability to adapt its noise level according to social context.
% Four participants rated the noise level the same for both gaits as 3: acceptable, and all the other participants rated MPC as higher noise level than RL. \sehoon{More direct stats, followed by intuition}
% Although the rates were consistent, the preference on the noise level was diverged. \sehoon{How do we know preference only regarding the noise? interview?}
% One group preferred a quiet robot to avoid drawing attention disturbing others, especially in quiet environments like offices.
% Notably, participants with hearing impairments all favored the robot to make minimal sound as possible to concentrate on surroundings, with P9 mentioning a sense of increased safety during the trial with the lower noise.
% The other group favored a robot that makes some sound to indicate its operational status and location, which would be particularly useful in noisy environments such as stations or shopping  malls to reduce mental effort and the chance of losing track of the robot.
% For those desiring a consistent noise from robot, they expected the robot's noise level to adjust rationally in response to the environmental noise, louder robot sounds in noisier surroundings.
% \sehoon{This is a very counter-intuitive- not all prefer silence. We should stress it.}

\mypara{Feel reliable with less compliant RL gaits}
% Among all the participants, 
Nine out of 18 participants indicated that the MPC gait (average: 3.00) is more compliant than the RL-based gait (average: 2.44). This perception is because MPC's higher foot clearance typically leads to wider footsteps upon external perturbation. 
% We also noticed that some users still rated the MPC policy as more resistive, even if they accurately described the behaviors of each gait. This confusion arose because the robot required larger navigation actions to correct its deviated trajectories. In this case, we clarified the users with the intended definitions. 
% \sehoon{I deleted the paragraph about confusion. We should manually correct them in our data.}

% Participants showed varied responses to the robot's resistance level, despite the resistance level was clearly set and participants could discern the differences (average rating difference; 1.89 \sehoon{start with more direct stats}
% This variation is partly attributed to the participants' unfamiliarity with robotics terms; for instance, 'resistance' was often perceived negatively. 
% One participant associated less resistance with less conflict, preferring the RL policy, despite its higher compliance.

% It was observed that while some participants accurately described the behavior of each policy, they still rated the MPC policy as more resistive. 
% This perception could be because that during the run with MPC policy, the robot often corrected its path, when deviated due to human force exerted through the handle. \sehoon{not clear. need Taery's explanation.}
% % easily moved by forces
% % on a pre-planned route in a wizard-of-oz setup
% This extra corrections led to MPC policy being perceived as more resistive particularly when the human handler made frequent or dynamic movements. \sehoon{Interesting. Is this true?}

Among 12 participants who showed a preference for the compliance level, nine preferred the gait to be less compliant because they felt that higher compliance diminished the reliability of the robot as a guide.
% \sehoon{The other nine preferred more compliant gaits? Or nine preferred RL due to their compliance} 
\user{3} commented that the MPC gait felt ``too loose, felt lost'' and found that the RL gait ``felt more confident.''
Moreover, increased compliance often caused the robot to adjust its position more frequently, leading to more collisions with the user. While a few considered contact with the robot to be helpful for locating the robot, the majority of users felt less comfortable with such collisions.

% Two main factors influenced participants' preferences regarding resistance: the incidence of human-robot collisions and the perceived reliability of the robot. 
% Higher compliance tended to result in more collisions, especially during turns, but reactions to these collisions were mixed. 
% Some participants found them natural and helpful indications for locating the robot, while others felt less comfortable.
% Also, some felt that higher compliance made the robot seem less reliable and more pet-like than guide-like. 
% Notably, participants with imbalanced gaits preferred higher resistance, perceiving it as a more trustworthy and robust policy.

\mypara{Satisfaction and design implication}
In combination, the participants preferred the RL gait (average: 4.59) over the MPC gait (average 3.75) as illustrated in Fig.~\ref{fig:q1-gait}. Out of 12 participants who preferred the RL, four participants preferred the RL due to their reduced noise level, five participants reported their less compliant behaviors as the reason, and others mentioned overall impressions, such as being easy to walk with. 
% \sehoon{what about the other three?} 
Still, \user{10}, \user{13}, and \user{15}  liked the MPC gait because it allowed them to stop or redirect the robot when needed, for cases such as pulling back the robot from tugging forward or stopping when a door is closing. Therefore, it would be best to offer customization of both noise and resistance levels depending on user preference and social contexts.

\vspace{-0.2em}
\section{Part 2. Navigation Movements} \label{sec:exp2-human-movement}
\vspace{-0.45em}
Guide robots make diverse navigation movements 
% to avoid obstacles and 
to lead humans, and previous studies aimed to model humans following robots~\cite{nanavati2019follow,wang2021navdog,xiao2021robotic,morlando2023tethering,kim2023transforming}.
While some movements are carried out for efficiency, they may not always align with the comfort and expectations of the user.
This part of the study aimed to understand a human's preference for the robot's navigation movements, such as turns, sidesteps, and backsteps.
% , which a robotic guide dog may employ during navigation.
We looked at the impact of these movements on human handlers by assessing clarity, safety, and frustration levels.

\vspace{-.35em}
\subsection{Procedure} \label{sec:exp2-human-movement-procedure}
\vspace{-.35em}
\mypara{Movement types}
% \sehoon{A small figure can be useful here.}\taery{-appendix?}
We categorized the robot's movements into turns and directional steps.
The turns were further differentiated based on the direction (left or right), mechanism (robot-centric rotating in-place or gradual curve), and speed (slow or fast).
In-place turns involved the robot remaining stationary with no forward momentum during turning.
The robot slowed down to half the standard forward speed for slow gradual turns, while it maintained the standard forward speed for fast gradual turns.
The standard forward speed was customized to each user during the initial familiarization phase.
Directional steps included movements to the left, right, and backsteps.
A diagram of these movement types can be found in the appendix.

% in-place slow: fwd 0, turn 
% in-place fast: fwd 0, turn*2
% gradual slow: fwd*.5, turn
% gradual fast: fwd, turn*2

% left
% in-place slow
% in-place fast
% gradual slow
% gradual fast

% right
% in-place slow: backstepping
% in-place fast: backstepping
% gradual slow: human stop, inplace
% gradual fast: human gradual

\mypara{Study procedure}
To mimic the unpredictability typically encountered in real-world navigation, the robot moved forward for a few seconds before executing each type of movement.
Participants experienced each movement type three times and assessed each based on clarity, security, and frustration level using a 5-point Likert scale.
\textit{Clarity} was a measure of the ease of recognizing the guide's intended movement.
\textit{Security} measured the perceived safety of the movement. And \textit{frustration} gauged any feelings of disappointment or annoyance that might arise~\cite{hart2006nasa}.
Additionally, participants were asked to indicate their most and least preferred turn types for each group.
The order in which participants tried out left and right turn types, as well as the type of locomotion, MPC or RL, was counterbalanced.

\vspace{-.5em}
\subsection{Results}
\vspace{-.25em}
\begin{table*}[ht]
\caption{
Comparative Ratings and Votes for Turn and Step Types.
Participant ratings (1 to 5) for clarity, safety, and frustration of in-place~(IP) and gradual~(G) turns~(differentiated by direction and speed), sidesteps, and backward steps, with standard deviations in parentheses.
Votes for the most and least preferred for each type are also included.
% Note that we excluded votes if 
}
\centering
\scriptsize
\vspace{-1em}
\setlength\tabcolsep{6.5pt}
\begin{tabular}{@{}lccccccccccc@{}}
\toprule
\textbf{} & \multicolumn{4}{c}{Left Turn} & \multicolumn{4}{c}{Right Turn} & \multicolumn{3}{c}{Step} \\ \midrule
\multicolumn{1}{l|}{\textbf{}} & IP-Slow & IP-Fast & G-Slow & \multicolumn{1}{c|}{G-Fast} & IP-Slow & IP-Fast & G-Slow & \multicolumn{1}{c|}{G-Fast} & Left & Right & Back \\ \midrule
\multicolumn{1}{l|}{\textbf{Clarity} $\uparrow$} & 4.3 (1.0) & 4.6 (0.9) & \textbf{4.8 (0.4)} & \multicolumn{1}{c|}{4.5 (0.8)} & 4.1 (0.9) & 4.3 (0.8) & \textbf{4.7 (0.5)} & \multicolumn{1}{c|}{4.7 (0.6)} & 4.1 (0.9) & 3.6 (1.2) & \textbf{4.6 (0.8)} \\
\multicolumn{1}{l|}{\textbf{Security} $\uparrow$} & 4.4 (0.8) & 4.2 (1.1) & \textbf{4.9 (0.3)} & \multicolumn{1}{c|}{4.5 (0.8)} & 4.2 (0.9) & 4.1 (1.1) & \textbf{4.8 (0.5)} & \multicolumn{1}{c|}{4.5 (0.7)} & 4.3 (0.9) & 3.9 (1.2) & \textbf{4.7 (0.8)} \\
\multicolumn{1}{l|}{\textbf{Frustration} $\downarrow$} & 1.9 (1.3) & 1.6 (0.8) & \textbf{1.1 (0.3)} & \multicolumn{1}{c|}{1.3 (0.6)} & 1.9 (1.4) & 1.9 (1.3) & \textbf{1.2 (0.6)} & \multicolumn{1}{c|}{\textbf{1.2 (0.5)}} & 1.5 (0.8) & 2.0 (1.4) & \textbf{1.2 (0.5)} \\ \midrule
\multicolumn{1}{l|}{\textbf{Best}} & 1 & 2 & \textbf{7} & \multicolumn{1}{c|}{5} & 0 & 3 & 5 & \multicolumn{1}{c|}{\textbf{9}} & - & - & - \\
\multicolumn{1}{l|}{\textbf{Worst}} & 6 & 2 & \textbf{0} & \multicolumn{1}{c|}{2} & 6 & 2 & \textbf{1} & \multicolumn{1}{c|}{\textbf{1}} & - & - & - \\ \bottomrule
\end{tabular}
\vspace{-2.5em}
\label{tab:q2-movement}
\end{table*}
Table~\ref{tab:q2-movement} presents the survey result, including average ratings and number of votes received.
The total votes do not sum up to the number of participants (n=18) because some participants expressed no preference for specific movements.
To analyze the ratings, we employed the Friedman test, followed by pairwise post-hoc comparisons using the Nemenyi test.
Additionally, the Spearman's rank correlation test was conducted to find potential correlation between the votes and ratings. 

Overall, the asymmetry of the human-robot team, in which a handler always holds a dog with their left hand, caused interesting preference differences between left and right movements.
% \sehoon{We may need a small figure for illustrating asymmetric preference}

\mypara{Gradual slow movements preferred for left turns}
Our analysis using the Friedman test revealed significant differences in security and frustration levels among the left turn types ($p < 0.05$).
% We further examine the significant pairs through a Nemenyi all-pairs test.
% (safety ip-f, g-slow)
% \sehoon{I would start with the best option (G-slow, I guess) that would draw attention from readers.}
The gradual-slow left turn was deemed safest, allowing handlers to maintain or slightly reduce their walking speed during the turn.
On the contrary, the in-place-fast turn is perceived as least safe,  due to participant unease, as expressed by  ``[I] felt it was going to run me over'' (\user{11}) and ``[I] had to get out of its way'' (\user{18}).
This is because the fast left turns did not give participants, who were located on the right side of the robot, sufficient time to adjust their walking.
% While there were frequent collisions observed and also mentioned by subjects (P2, P5) during in-place-slow left turn, they rated in-place-fast left turn to be less safe,
% ip-s, safety: ``because it was slow, it banged the leg'', ``foot kept hitting the dog'' pace didn't match
% Because participants were walking on the robot's right side, they needed more movement than the robot during left turns, and the fast turns did not give them sufficient time to adjust their walking patterns.
% On the contrary, the gradual-slow left turn was deemed safest, allowing handlers to maintain or slightly reduce their walking speed during the turn.
% % ip-f: P17 very safe, very clear, but frustrating-didn't like it wasn't smooth)
% (frustration ip-s, g-slow)

In terms of frustration level, in-place-slow turn scored lowest whereas gradual-slow scored highest; however this result did not reach conventional levels of statistical significance ($p=0.18$).
Participants mainly discussed the speed and smoothness when rating the frustration level.
For instance, \user{03} found the in-place-slow turn very frustrating (rated 5), mentioning ``too slow'', and \user{17} disliked its ``abrupt 90 degree turns''. % ``didn't like it (in-place-slow left) was making abrupt 90 degree turns''

% vote: 
Participants generally preferred gradual turns, with seven votes for gradual-slow and five for gradual-fast, appreciated for their smooth pacing and natural feel.
This preference was negatively correlated with frustration ($-0.356$, $p=0.002$), with several participants noting that these movements provided a sense of ease and were similar to the natural movement of a dog. 
Conversely, the least preferred was the in-place slow turn, with participants reporting a dislike for the robot's stopping (\user{07}), its slow pace (\user{09}, \user{13}), and the turn's unexpected nature (\user{14}, \user{17}), which required participants to move out of the way of the rotating robot.
Clarity was not significant for voting the most preferred turn (0.166, $p=0.16$), but it was for the least preferred ($0.373$, $p=0.0012$). 

Interestingly, all votes favoring the gradual-fast left turns came from current or previous guide dog handlers, commenting ``it was smoother and faster'' (\user{14}).
However, this same type of turn was least preferred by two cane users with high frustration rates, possibly due to its relatively hasty movement for those accustomed to slower walking.
% P9 moving a little bit faster
% because of their preference and experience for speed.
Gradual-fast turns were the fastest among all types of turns, but guide dog users had ample experience with the faster gait during left turns when the guide is on their left. 
This suggests that previous experience with guide dogs influences speed and style preferences.

\mypara{Gradual fast movements preferred for right turns}
One notable difference for right turns is a preference for a gradual fast turn, which was selected as the most preferable turn with nine votes. 
Participants frequently cited ``naturalness'' as the main reason for their preference, specifically pointing to the ``pace'' of the movement. 
\user{10} highlighted that the ``gait matched better (turning radius and pace),'' which made the movement feel more natural. 
On the other hand, the gradual-slow right turn was less favored despite the highest rating.
As a result, the best turn selection does not show statistical significance with clarity, security, and frustration ratings ($p>0.1$), while the worst turn selection does ($p<0.05$).

Similarly to left turn scenarios, in-place right turns were generally rated more negatively than gradual turns, mainly because they required handlers to backstep, impacting all three metrics. \user{02} and \user{17} noted that in-place turns involved excessive turning and felt counterintuitive, affecting both clarity and frustration levels. 
Conversely, \user{03} described gradual turns as natural and comparable to a guide dog's movements. 
Safety concerns were highlighted by \user{18}, who had to move backward to avoid human-robot collisions.

Still, three participants exhibited a preference for in-place fast turns despite the general inclination toward gradual turns. 
These individuals all shared a slower walking speed that they preferred, and they appreciated the slower-paced movements.
% \sehoon{The prev sentence is not clear}
% Also, they are all trying with the MPC gait. 
% This gait's high compliance to external forces likely resulted in less required backstepping, enhancing their comfort with the movement. 
% \sehoon{Not sure readers can follow the previous sentence}
% When voting for the least preferred turns, the participants expressed significant concerns about safety (0.427, $p=0.00018$), particularly repeating issues related to backstepping and potential human-robot collisions.

\mypara{Sidesteps and backsteps}
Participants generally rated left sidestepping more positively than right sidestepping.
Because the latter moves toward the handler,
% \sehoon{that is moving toward the handler...?}
the right sidestep received the lowest ratings of all movements, impacting all three metrics. 
\user{03} and \user{16} noted that the sidestepping movement was not like that of a dog, which would normally move around slowly, while \user{15} mentioned that both sidesteps were a bit of a surprise.
\user{02} and \user{17} found the right sidestep especially unclear, causing increased frustration, as it lacked the instantaneous pulling force that occurs in the left sidestep.
For security, participants felt the robot would sense the environment when sidestepping to the left, whereas moving right put the human handler in a more vulnerable position.
\user{18} expressed concern that they ``might have something on the right'' and thus did not feel completely safe.
Additionally, because the robot leads the handler, rightward movements possibly result in human-robot collisions (\user{07}, \user{17}).
% Participants expressed concern about the possibility even in the absence of a collision (\user{07}, \user{17}).

When participants were informed that back stepping would be included, they expressed initial concern, but frustration ratings after trials were as low as those for right turns.
\user{17} found it very clear, especially due to the slight stop in forward momentum, which felt natural when transitioning from forward to backward movement.
Although \user{18} stayed uncomfortable about backstepping, citing safety concerns, \user{03} and \user{16} indicated that backstepping was acceptable, even though it does not happen frequently for actual guide dogs.

Given that sidestepping and backstepping are expected to occur less frequently, \user{03} and \user{15} suggested adding extra signals, such as beeps,
% beeps for sidestepping and a long beep for backstepping, 
to improve clarity and user comfort.

\vspace{-0.5em}
\section{Part 3. Bidirectional Communication}\label{sec:p3-bi-comm}
\vspace{-0.5em}
Guide dogs and users need bidirectional communication for effective navigation as a team.
Bidirectional communication includes nonverbal interactions via handles, such as a harness or a leash, and verbal instructions from users.
We first examine two different handles: a harness and a leash. While a harness has been considered as a major option, several research efforts investigated a leash as an alternative~\cite{xiao2021robotic, chen2023quadruped, defazio2023seeing, morlando2023tethering}.
We also explored different verbal commands inspired by both practices with an animal guide dog and autonomous navigation research.
% Guide dogs are typically equipped with both harness handles and leashes to guide humans. 
% Similarly, leashes are used in several guide robot research efforts using legged robots\cite{xiao2021robotic, chen2023quadruped, defazio2023seeing, morlando2023tethering}.
% On the other hand, human handlers give some verbal instructions to initiate travel with guide dogs. 
% In this part, we explore the following regarding the bi-directional communication:
% \begin{itemize}
%     \item Which type of handle is preferred? How do flexible leashes and rigid harness handles differ in robot-to-human communication? What should be considered when designing a handle to have an effective guide from robot to human?
%     \item What instructions do human handlers want to give to guide robots for navigation?
% \end{itemize}

\vspace{-.5em}
\subsection{Procedure}
\vspace{-.5em}
\mypara{Robot to human}
Because participants already experienced interactions with a harness, we asked users to experience a leash only for this time.
% Following the previous studies, we switched the handle to a flexible leash.
Participants were asked to walk with the robot in the lab space for three to five minutes, as done in Part 2 
% The robot performed random movements, including turns, sidesteps, and backsteps, as done in Part 2 
(Sec.~\ref{sec:exp2-human-movement-procedure}).
Participants then rated clarity, security, and frustration on a scale of -2 (decreased experience) to +2 (increased experience),
% \sehoon{Is this matched with Part 2? Maybe better to have 1 to 5 for consistency}
comparing their overall experience using the flexible leash to their experience with the harness.

\mypara{Human to robot}
We provided a list of possible instruction types, and participants were asked to rate their expected usefulness.
Unlike other parts of the study that used a wizard-of-oz approach, this subpart relied on verbal interviews instead of direct interactions.
On top of directional cues that are popular in practice, we also suggested the instructions associated with objects, areas, final destinations, and point goals, which are known as common navigation tasks suggested by Anderson et al.~\cite{anderson2018evaluation}. 
Each instruction was rated on a scale of one (not useful) to five (very useful), and participants brainstormed additional tasks they expected the guide robot to perform.

\vspace{-0.5em}
\subsection{Results}
\vspace{-0.5em}
% \subsubsection{Handle type}
\begin{figure}
    \centering
    \includegraphics[width=.85\linewidth,trim={0 4.6cm 0 10cm}, clip]{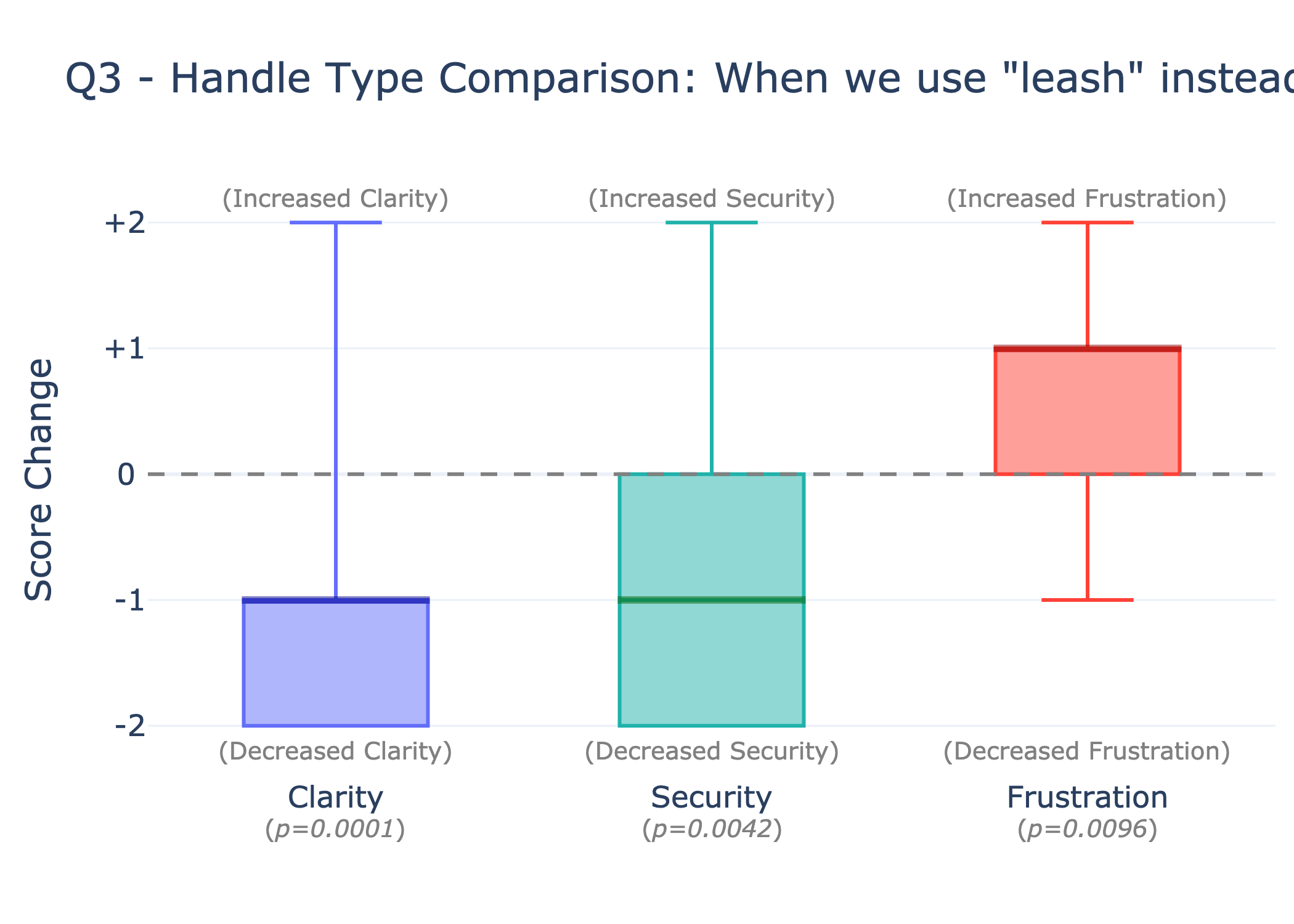}
    \vspace{-1em}
    \caption{
    % Score change when switched to leash handle from rigid handle. 
    Ratings of using leash handles on clarity$\uparrow$, security$\uparrow$, and frustration$\downarrow$, compared to rigid handles on a scale of -2 to +2.
    Overall, leash handles were rated lower in clarity and security and higher in frustration than rigid handles.
    % change to left-right, decreased == rigid handle; increased == leash handle; 0 == similar experience
    }
    \label{fig:q3-1-leash}
    \vspace{-1.75em}
\end{figure}

\mypara{Preference for a harness over a leash}
Overall, 16 out of 18 participants preferred the rigid harness handle for guiding due to clearer information and reduced mental demand.
Fig.~\ref{fig:q3-1-leash} indicates significant decreases in clarity and security and an increase in frustration when switching from a rigid harness handle to a flexible leash ($p<0.01$).
\user{03}, \user{13}, and \user{17} emphasized the clarity and certainty provided by the harness handle.
The leash led to more mental demand and less relaxation due to its less clear feedback and increased safety concerns, making it hard to determine the status of the robot.

Two other participants preferred a leash for its extra space and reduced human-robot collisions.
Still, they noted the issues with prompt reactions and orientation matching.
% P8 and P11 found the leash provided more space but required more effort to interpret movements.

\mypara{Ambiguous communication via a leash}
Most participants found the leash less clear, requiring them to hold it tightly to maintain tension or even shorten the leash.
\user{07} specifically noted that clarity mainly decreased for right turns, while turning left was less problematic because it pulls the leash.
There was a delay in identifying the robot's movements when the leash was slack. 
As a result, participants often paused until they were sure about the robot's heading.

Despite the decrease in clarity, participants managed to follow the guide by using other senses with increased mental demand.
% P9 commented on using the robot's sound and vision to follow, indicating higher mental effort. [any other participants?]
They relied on their remaining vision (if any), sound from the robot, and keeping the leash taut.
For instance, during backstepping, they sensed that the robot was approaching closer to them, indicating that it was backing up.

Although not all participants noticed, we observed confusion between turns and sidesteps, which caused them to misinterpret sidesteps as turns.
Unlike the rigid harness, the leash did not provide an immediate and definite guidance.
Participants moved towards the source of the force, facing the indicated direction. 
As a result, sidesteps were often misinterpreted as turns. 
% P18 noted difficulty in distinguishing sidesteps from turns.
% P ?? mentioned that there were subtle difference, but they didn't even know that sidestepps were happening.
This suggests the robot should plan routes differently when using a leash, as the guide information was less certain.

\mypara{Consideration for equipping both handle types}
We also asked if using both types of handles would benefit users,  based on the fact that many guide dogs wear both for different roles.
Fifteen participants agreed on using both types.
Guide dog users favored familiarity, suggesting the same usage as animal guides, while others considered the benefits of using a leash for less important guiding, such as in a familiar environment (\user{08} and \user{11}).
However, \user{03}, \user{13}, and \user{18} preferred not to have a leash, as they found it unhelpful.
% P3 leash guide training: holding leash short. 
% Rigid handle offered more control and favored in crowded environments

\begin{table}[ht]
\vspace{-1.5em}
\centering
\caption{Usefulness rating on human-to-robot instruction types
% \sehoon{highlight some interesting numbers}
}
\vspace{-1em}
\scriptsize
\begin{tabular}{@{}lcccc@{}}
\toprule
\multicolumn{1}{c}{\textbf{Instruction Type}} & \textbf{Mean} & \textbf{Std} & \textbf{Min} & \textbf{Max} \\ \midrule
Direction (forward, left, stop) & 4.9 & 0.2 & 4 & 5 \\
Object (chair, sofa, table) & 4.9 & 0.2 & 4 & 5 \\
Area (restroom, office, lobby) & \textbf{5.0} & \textbf{0.0} & 5 & 5 \\
Final destination (library, bus stop) & 4.7 & 0.7 & 3 & 5 \\
Point goal (15 feet North) & \textbf{3.0} & \textbf{1.7} & 1 & 5 \\ \bottomrule
\end{tabular}
\label{tab:3-2-instruction}
\vspace{-0.5em}
\end{table}
% \subsubsection{Navigation Instruction}
\mypara{Basic expectation for human-to-robot instructions}
Participants rated the usefulness of different verbal instructions as shown in Table~\ref{tab:3-2-instruction}.
The results indicate high ratings for most navigation goals, with a few exceptions. 
First, participants expressed their expectations for guide robots' interactions to be at least at the guide dog level.
This expectation is reflected in the high rating of 4.9 for directional goals, which are common verbal cues for guide dogs.
Participants often expected direction goals to include detailed commands, such as \textit{over left/right} for sidestepping (\user{14}), to maintain the guidance quality.
% These ratings highlight the perceived usefulness for various instructions and provide insights into potential user preferences and needs.

% First, participants expressed their expectations for guide robots' interactions to be at least at the dog-guide level.
% For instance, direction goals received an average rating of 4.9, which are often considered as basic types of commands for guide dogs.
% Participants often expected direction goals to include detailed commands such as \textit{over left/right} for sidestepping (P14).

% Direction goals and object goals received an average rating of 4.9.
% These instructions are similar to those used with guide dogs, reflecting participants' expectations for interactions at least at the dog-guide level.
% For example, participants expected direction goals to include detailed commands such as \textit{over left/right} for sidestepping (P14).
\mypara{Human-to-robot communication beyond animal guides}
The key potential of robot guide dogs is to understand extended user commands by leveraging modern natural language processing technology.
Object goals were expected to exceed current guide dog capabilities, with specific instructions like \textit{front door of the car} or personalized knowledge such as \textit{my dad's car} (\user{12}).
Area goals are also known as AreaNav in the navigation community, which aims to move to specific areas, such as \textit{restroom or office}.
Both object and area goals received the high ratings of 4.9 and 5.0.

The goal of giving a final destination received an average of 4.7, which was not rated as highly as area navigation (5.0) and not even direction goals (4.9).
Despite the convenience of specifying a final destination, participants expressed concerns about potential miscommunication (\user{04}, \user{06}, \user{18}).
They worried that the robot might take them to a different location than intended, especially in cases of places with similar names (e.g., two different Starbucks). % requires additional interaction to confirm
Participants preferred using more detailed cues for more certainty. % rather than getting lost

Point goal navigation, which sets the goal using distance and cardinal directions, is the most common navigation problem.
It received varied ratings (ranging from 1 to 5, with a median of 3).
Seven participants found it useful for following directions from a mobile navigation app or when someone explains the way in such a manner.

% \mypara{Summary}
% Overall, the expected level of communication was similar to that with guide dogs but with technological enhancements.
% Participants suggested other potential instructions to improve guidance, such as asking when to cross streets, finding railings, and following someone.
% They also mentioned synchronizing the robot with smartphone apps and using technological intelligence to identify objects and people in their environment. \sehoon{We may delete this paragraph if we are running out of pages}
\vspace{-0.5em}
\section{Part 4. Explainability}\label{sec:p4-explainability}
\vspace{-0.5em}
% Explainability in human-robot interaction refers to the robot’s ability to explain its behavior. It helps the human user to understand the mental model of the robot so that they can have reasonable expectations of the robot’s behavior and limitations and can trust its actions \cite{Setchi2020}, \cite{Dehkordi2021}. Please see the survey paper \cite{Anjomshoae2019} to see a detailed description of current research. Our work seeks to determine the benefits of audio explanations for improving the user experience when using a robotic guide dog.

We explore the potential explainability feature, which refers to the robot’s ability to reason about its behavior~\cite{Anjomshoae2019}.
Explainability can be a key feature provided by a guide robot, unlike an animal guide.
% Furthermore, the guide robot can carry its entire system, while a smart cane requires the user to carry the computer and batteries.
% One of the reasons that white cane users like their mobility aid more than guide dogs is that a white cane helps them identify obstacles, while a guide dog does not provide such information.
Inspired by that, this section examines whether the explainability of guide robots can improve the user experience, and which features contribute the most.
% We explore diverse sets of information that could help the users navigate in a building.

% include obstacle information.
% Also, some other information that could help the users navigate and extra information

\vspace{-0.7em}
\subsection{Procedure}
\vspace{-0.3em}
% \textbf{Explanation Audio.}
We designed our experiments to explore a diverse set of verbal cues given by audio that can help the users navigate in a building.
Our navigation task was designed as a round trip, consisting of \textit{``finding a restroom''} and \textit{``returning to the room''}.
The experiments began with the participants' verbal cue, and the robot provided audio explanations during autonomous guidance, such as navigation assistance (e.g., \textit{``we are leaving the room''}), safety advisories (e.g., \textit{``there is a wet floor''}), environmental context (e.g., \textit{``we are passing by Room 018''}), or additional information (e.g., \textit{``the weather looks good''}).
We synthesized audio using Google Cloud TTS.
% We prepared audio for two navigation tasks: \textit{``Find the restroom''} and \textit{``Take me back''} for a round trip.
% The participants gave verbal instructions to the robot to start their navigation.
The audio explanations informed the next movement, obstacles, and surroundings.
% We used Google Cloud TTS to generate the audio.
The participants traversed the path one way with audio and the other without audio explanation; the order was counterbalanced for each participant.

Then, we asked 11 (Table~\ref{tab:q4-1-piads}) questions to examine the impact of the guide robot with or without audio explanation.
The questions were developed based on the Psychosocial Impact of Assistive Devices
Scale (PIADS)~\cite{jutai2002psychosocial} with one additional question on \textit{trust} (i09) in mobility aids from a modified PIADS for guide robots~\cite{zhang2023follower}.
Participants were asked to compare the two settings of the guide robot (with and without audio explanation) against their current mobility aid, answering on a scale ranging from -2 to +2, where -2 indicated ``much worse'', +2 indicated ``much better''.
Also, we reviewed the phrases to assess and understand the usefulness of each phrase in the audio explanation, using a Likert scale from 1 to 5, where 1 indicated ``not useful at all'' and 5 indicated ``very useful.''

\vspace{-0.5em}
\subsection{Results}
\vspace{-0.5em}
\begin{table}
\centering
% \scriptsize
    \setlength\tabcolsep{8pt}
    \caption{Modified PIADS items to compare robotic guide dog experience to user's current mobility aid (Descriptions in appendix)
}
\scriptsize

\vspace{-1em}
\begin{tabular}{cc}
\toprule
\textbf{ID} & \textbf{Item} \\ \midrule
i01 & Ability to adapt to the activities of daily living \\
i02 & Ability to participate \\
i03 & Ability to take advantage of opportunities \\
i04 & Eagerness to try new things \\
i05 & Happiness \\
i06 & Independence \\
i07 & Productivity \\
i08 & Quality of life \\
i09 & Trust \\
i10 & Security \\
i11 & Embarrassment \\ \bottomrule
\end{tabular}
\label{tab:q4-1-piads-short}
\vspace{-1.em}
\end{table}
% \vspace{-.25em}
\begin{figure}
    \centering
    \vspace{-.5em}
    \includegraphics[width=1.0\linewidth,trim={0 0 0 .25em}, clip]{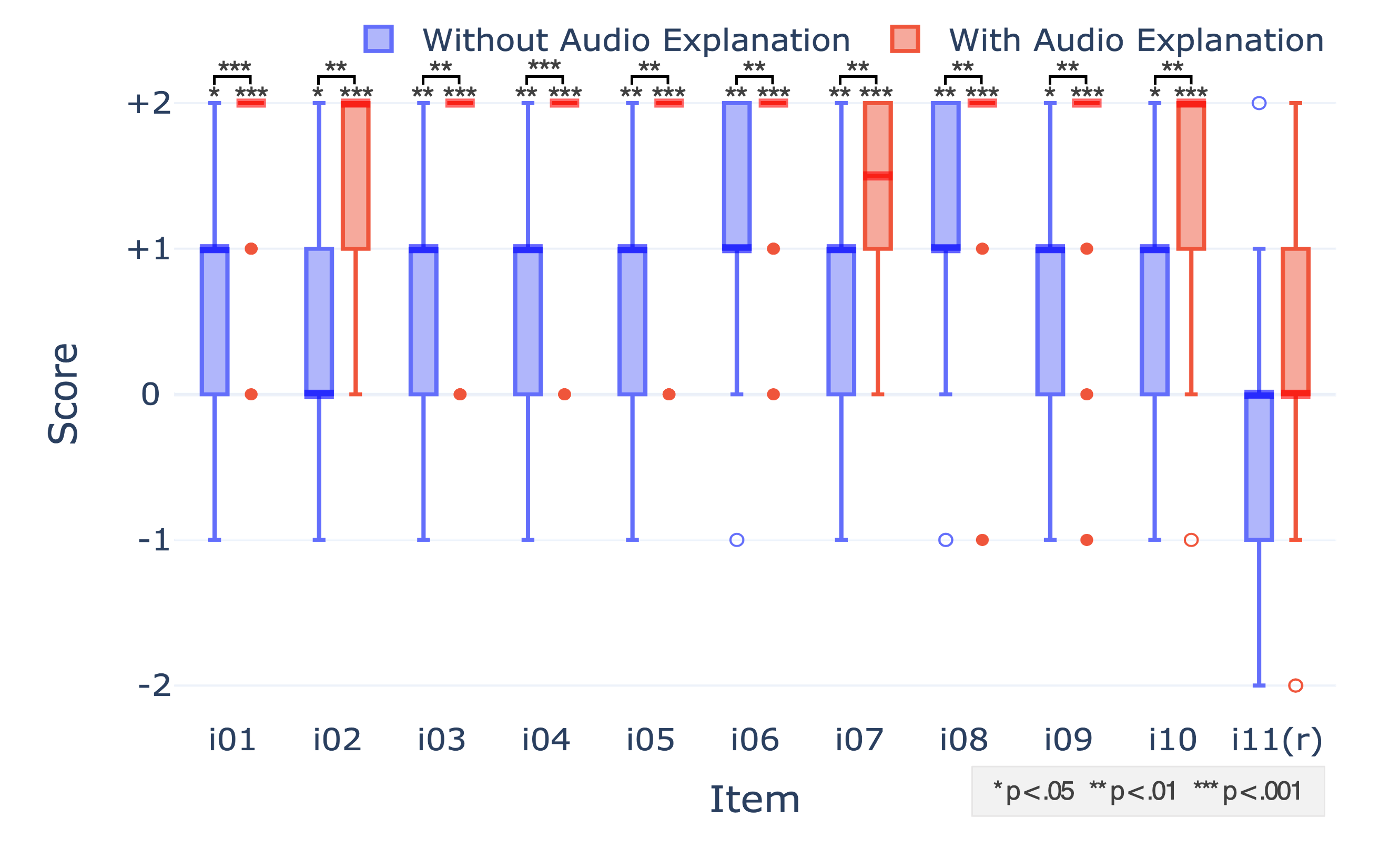}
    \vspace{-2.5em}
    \caption{Ratings of Modified PIADS (n=14) for the robotic guide dog without and with audio explanation. Overall, participants estimated a positive impact from the robotic guide dog alone, and greater when audio explanation is included, compared to their current mobility aids. Item i11 reverse coded.
    % \todo{add *, reverse code i11?}
    % \sehoon{add a key message, something like clear preference of audio explanation over...}
    }
    \label{fig:q4-1-audio}
    \vspace{-1.5em}    
\end{figure}

\mypara{Enhanced user experience}
Fig.~\ref{fig:q4-1-audio} highlights significant differences when comparing participants' experiences with their current mobility aids to their experiences with the guide robot, both with and without audio explanations.
% Wilcoxon signed-rank test
These results show a clear preference for the guide robot, particularly when audio explanations are provided.
On average, the audio explanation improved user experience by +1.48 (std=0.48), which is greater than without audio of +0.67 (std=0.31).
% , when compared to their current mobility aid on a scale ranging from -2 (much worse) to +2 (much better).
% \sehoon{Any statistical summary? e.g., audio explanation improved XX, which is bigger than without audio of YY...}

Participants noted that the guide robot with audio made them feel more interactive with their surroundings and the robot.
\user{05} mentioned it would become a ``conversational piece,'' enhancing self-esteem and promoting independence.
\user{02} said the experience was ``not walking to be walking but it was \textit{participating} - it was \textit{alive}.''
\user{07} found it easier than using a white cane, and \user{15} felt more self-confident.
\user{09} highlighted that the audio led to a safer experience by balancing their stride more effectively.
\user{18} found the robot with audio ``much better than the white cane,'' appreciating the additional feedback.
In contrast, one user, \user{17},  expressed a level of distrust, noting, ``I'm used to getting information from the cane, haven't built a trusting relationship with the robot,'' illustrating the challenge of shifting from traditional aids to technology-based solutions.
% In contrast, P18 found the robot with audio ``much better than the white cane,'' appreciating the additional feedback.

\mypara{Minor concerns over privacy and embarrassment}
While most responses were positive, the responses to embarrassment (i11) revealed diverse opinions. 
Six participants felt no difference in embarrassment between using their current aid and the guide robot, with or without audio.
Yet, \user{13} and \user{15} noted potential embarrassment due to the robot's appearance and noise. 
Conversely, \user{07} and \user{11} viewed the robot as ``cool technology'' and thought it might rather reduce embarrassment compared to their current mobility aids.
\user{18} and others raised concerns about privacy with the audio output, suggesting that earbuds could mitigate embarrassment by keeping the conversation private. 
This mix of feedback underscores the importance of customizable and discreet communication options in enhancing user experience without drawing unnecessary attention.

% Also, the guide robot with audio explanations was seen as a significant improvement in making navigation easier and less embarrassing (P5, P10), particularly when it helped prevent participants from getting lost during turns, as P10 highlighted.
% This supports the need for explainable features to enhance the clarity and safety of navigation with guide robots.
% \sehoon{This paragraph is a bit out of story- maybe merged to the second paragraph.}

\begin{table*}[ht]
\setlength\tabcolsep{5pt}
\centering
\caption{Phrase examples used for explainability and usefulness ratings from participants (n=18) on a 5-point Likert scale}
\scriptsize
\vspace{-1em}
\begin{tabular}{llccccc}
\toprule
\multicolumn{1}{c}{\textbf{Category}} & \multicolumn{1}{c}{\textbf{Phrase Example}} & \textbf{Mean} & \textbf{Std} & \textbf{Med}& \textbf{Min}& \textbf{Max} \\ \midrule
\textbf{Navigational Assistance} & Directions (e.g., ``Going forward.") & 4.6 & 1.14 & 5 & 1 & 5 \\
\textbf{} & Space Change (e.g., ``We are now leaving the room and entering a hallway.") & {4.9} & {0.32} & 5 & 4 & 5 \\ \midrule
\textbf{Safety and Caution Advisories} & Obstacle (e.g., ``We are avoiding an obstacle on our left, possibly a trashcan.") & 5.0 & 0.00 & 5 & 5 & 5 \\
\textbf{} & Warning - Narrow Path (e.g., ``This path is narrow, be careful.") & 4.9 & 0.24 & 5 & 4 & 5 \\
\textbf{} & Warning - Wet Floor (e.g., ``There is a wet floor warning, be careful.") & 5.0 & 0.00 & 5 & 5 & 5 \\ \midrule
\textbf{Environmental Context} & Room (e.g., ``We are passing by Lab 018 on the right.") & 4.9 & 0.25 & 5 & 4 & 5 \\
\textbf{} & Exit (e.g., ``We are passing by an exit to the outside on the right.") & 5.0 & 0.00 & 5 & 5 & 5 \\
\textbf{} & Scene (e.g., ``You'll find tables, chairs, and people sitting around in this area.") & 4.7 & {0.85} & 5 & 2 & 5 \\
\textbf{} & Scene - Vending Machine (e.g., ``There is a vending machine on our left.") & 4.5 & 1.07 & 5 & 1 & 5 \\ \midrule
\textbf{Additional Information} & Acknowledgement (e.g., ``Okay, Let's go to the restroom.") & 4.8 & 0.51 & 5 & 3 & 5 \\
 & Weather Report (e.g., ``By the way, the weather looks good today!") & 4.3 & 1.41 & 5 & 1 & 5 \\ \bottomrule
\end{tabular}
\label{tab:q4-2-phrase}
\vspace{-2.75em}
\end{table*}
\mypara{Usefulness of all the phrases}
Overall, participants were positive about all the phrases, although there were minor concerns about some of them.
Participants rated navigation and safety phrases as highly useful, with a median score of 5 out of 5 across all types (Table~\ref{tab:q4-2-phrase})
% \sehoon{wasn't it -2 to 2?}.
Information such as detected obstacles, wet floor warnings, and exits received unanimous ratings of 5.
Both white cane and guide dog users appreciated this information as their current aids do not explicitly provide such details.
White cane users only receive limited feedback about obstacles and wet floors, often realizing the hazard too late, while guide dogs do not convey this information but take detours.
Participants emphasized the importance of alerts for terrain changes as these significantly enhance safety and convenience.
Clear indications of exits were also considered important information and essential for maintaining orientation and safety during navigation.

\mypara{Slight skepticism about directional and extra cues}
Information already communicated through physical interaction with the robot, such as directional cues, received mixed ratings.
While most participants found directional information very useful, commenting that it helped them prepare for turns and feel more relaxed, two participants rated it as low as 1. 
\user{18} felt such information was redundant as they could follow direction from the harness, and \user{17} noted that constant updates could become tiresome. 
Similarly, the narrow path warning message received a low rating from one participant who felt it was unnecessary as long as the robot successfully navigated the path.
% narrow path warning behavior: the participants tried to stay close to robot, sometimes behind the robot, to fit into the path.
% guide dog: narrow path -> slow down. --> still this feature is needed either via audio or behavior

Extra information not directly related to the navigation task, such as environmental context or weather reports, was deemed less useful.
Environmental context phrases, such as scene descriptions, received mixed feedback.
\user{18} suggested it would be useful in unknown environments but preferred to turn it off in familiar settings.
\user{13} found it useful only when searching for specific rooms, otherwise viewing it as unnecessary information.
Weather reports received the lowest mean rating of 4.3, which can be considered friendly small talk but less useful.
\vspace{-0.5em}
\section{Design Implications}
\vspace{-.5em}
% \mypara{Active involvement: control and surroundings awareness wanted}
%%SH: human to robot command
\mypara{Comprehensive high- and low-level commands for active involvement}
Throughout the study, participants desired comprehensive control over the robot.
Even with high-level autonomy available, they valued lower-level commands such as directional cues for more active navigation.
%%SH.7.29: In part 1, we said that users prefer less compliant gaits. This would be contradictory.
% Additionally, they wanted to make adjustments during navigation, favoring some level of compliance in robot gaits to adjust or correct the robot's movements.
We recommend future designers include a diverse set of user commands with different semantics, ranging from short-term goals (e.g., directions or objects) to long-term goals (e.g., final destinations) even within the same level of autonomy, as suggested in Part 3. 
This would also be followed by effective communication (Part 3) and explainability (Part 4), as clear awareness of the robot's status and environment was essential for effective control.
% , with some participants emphasizing the need for extra information to improve their mental maps.

% This need for control aligns with the findings from {zhang+ms} in that users preferred partial autonomy, demanding more control than full autonomy.
% zhang: i am the follower, also the boss.. ``Results showed that full autonomy received better walking performance and subjective ratings in the controlled study, whereas participants used more partial autonomy in the natural environment as demanding more control.''

 % (p1-preference on compliance; p3-high rate on direction cues).
% defazio2023seeing: force control

%%%%%%%%%%%%%%%%%%%%%%%%%%%%%%%%%%%%%%%%%%%%%%%%%%%%%%%%%%%

% \mypara{More information to relieve stress in navigation}
% \mypara{Information about the robot's status and environment is wanted}
\mypara{Reducing uncertainty to relieve stress during navigation}
Our studies revealed that stress during navigation was mainly due to uncertainties (Part 2).
For instance, movements toward the right cause frustration due to a lack of information on that side.
Participants looked for more environmental information through the robot, such as sensing the floor via the handle.
Explainability (Part 4) significantly increased the user experience by providing extra information, which would be an important factor for future robotic guide dogs.
% comparison to white cane (some info about obstacles) and animal dogs (limited information)
% robot status

Uncertainty can come from the robot itself, specifically the location of it.
% p1,p2,p3
For this reason, some participants prefer noisier gaits to locate the robot based on its sounds.
% Noise from the robot's gait and fan helped users locate the robot, with some preferring a noisier gait.
% , indicating that they keep look for such information.
% p2
Even gentle collisions with the robot were deemed as useful behavior that clearly tells the robot's location (Parts 1-2).
% wang2022can: ``I was particularly afraid that it stepped on my foot, I think stepping on me should be more painful than if the wheel rolled over me, anyway, I was very anxious.''
% p3, p4
Also, many users preferred a rigid handle because it reduced their mental load when locating the robot (Part 3).
From these findings, equipping a rigid handle and incorporating bidirectional voice interaction with explainability would be key measures to reduce user stress.

% p2 movement toward right - lack of information, (aligns with that ppl wanted the robot to be in front of them)
% p3 handle type: indirect. rigid handle help to feel the ground from the robot stepping.

%%%%%%%%%%%%%%%%%%%%%%%%%%%%%%%%%%%%%%%%%%%%%%%%%%%%%%%%%%%

%%SEHOON TODO: review this
\mypara{Adaptive behaviors for various users and situations}
% Tailoring to user and environment; history
% Of course, we want this, but for what?
% Not a single answer was the same!
Participants had varying preferences for gait and turning, indicating a need for more than just forward speed adjustments.
This includes turning radius and speed with adaptive movements, considering human's reactive movement to the guide (Part 2).
% These choices heavily rely on each person's walking habit, --> initial adjustment + as time goes by
% Requires to identify familiarity of environments (+ knowledge of personal historyrh)
% Familiarity with environments also influenced their expectations for handling choices and explainability.
% The utility of both handle types depended on the situation and familiarity with the environments.
In addition, the robots' behavior must consider environmental and social factors.
Participants wanted different levels of verbosity based on location familiarity, which results in different preferences over gaits, movements, handles, and explainability (Parts 1-4).
% Beyond the physical environment, the social environment also takes part in enabling personal queries related to the user's family or friends.
Therefore, preference learning as well as continual learning from history would be desirable.

% turning radius, turning speed, (e.g., applied to u-turn, avoiding obstacle. not just left/right turn for navigation)
% Variety in preference on gait and turning, not just forward speed adjustment is needed (p1, p2).
% p3 handle type switching-less guide needed, some prefer leash guiding. p3 shows that putting both types of handle would help, depending on the usage (mostly depending on familiarity with the location, but very rarely short-leash-guide is happening as per interview with Glacier+) + if handle type switched, different planning needed (p2)
% Identifying familiarity of environments (p4: different levels of explainability/verbosity expected based on familiarity with the location)
% Personal queries (p3-instruction: where is my dad's car; identify ppl)

%%%%%%%%%%%%%%%%%%%%%%%%%%%%%%%%%%%%%%%%%%%%%%%%%%%%%%%%%%%

%%SH: social aspects?
\mypara{Social acceptance for everyday life}
Participants expressed care about the people around them, aiming for a socially acceptable robot.
They didn't want to disrupt others with noise, suggesting adaptive noise levels for the robot's gait, and earbuds for hearing the robot's voice output.
They also emphasized following the pedestrian flow and disliked movements that could possibly disrupt it, such as abrupt turns.
Such behaviors require static or dynamic environmental information in real time to ensure that users can navigate smoothly and communicate effectively.
Participants expressed a willingness to engage in activities with others and viewed the robot as a potential conversational piece.
Therefore, robots should consider people around the user to blend into their lives naturally.
% (p1-noise control; p2-dislike abrupt turns, caring about pedestrian flow; p4-ability to participate)

%%%%%%%%%%%%%%%%%%%%%%%%%%%%%%%%%%%%%%%%%%%%%%%%%%%%%%%%%%%

\mypara{Participant Suggestions}
% Participants described several concerns and expectations regarding the use of guide robots.
% Operational reliability was a primary focus.
Participants emphasized the need for sufficient battery life, weather resistance for outdoor use, and manageable daily maintenance.
They suggest including active communication over these issues, such as low-battery warnings or self-checking reports. 
% The robot is expected to actively inform them about its state and give necessary information for easy maintenance.
% They also worried about the robot's reliability, questioning what would happen if it failed due to battery issues or malfunctions and what the user should do in such situations.
Safety and emergency response were also encouraged, such as calling 911 during an emergency.
% , with participants asking if the robot could detect emergencies and call 911 or ask for help.
% Concerns about theft and loss of the robot were also expressed.
Finally, affordability was stressed as an important factor in facilitating the widespread use of the guide robot.
\vspace{-0.5em}
\section{Conclusion}
\vspace{-.5em}
Our study explored various design parameters of robotic guide dogs to improve the user experience of blind and visually impaired individuals. Our studies examined control and interaction aspects, such as gait controllers, navigation behaviors, and interaction methods. By conducting user studies with 18 participants, the research identified various preferences over features, highlighting the need for customization and social awareness. We summarized those findings as design implications to provide essential insights for the future development of effective and user-friendly robotic guide dogs.

We have identified several directions for future research. Our study focused on two specific instances of locomotion controllers while leaving the development of a better controller as a future work. We also believe a harness or a leash can be improved for agile, non-verbal communication by offering joysticks or vibration feedback. Finally, we leave the visual appearance of robotic guide dogs unexplored, which would also have a significant impact on their social acceptance. 
It will also be important to discuss the psychological and social aspects of robotic guide dogs along with animal guides.

\vspace{-0.5em}
\section*{Acknowledgments}
\vspace{-0.5em}
% \vspace{}
We thank Jie Tan and Wenhao Yu for valuable discussions.
This research has been funded by the Industrial Technology Innovation Program (P0028404, development of a product-level humanoid mobile robot for medical assistance equipped with bidirectional customizable human-robot interaction, autonomous semantic navigation, and dual-arm complex manipulation capabilities using large-scale artificial intelligence models) of the Ministry of Industry, Trade and Energy of Korea.
JK was supported by the NSF GRFP under Grant No. DGE-2039655. Any opinion, findings, and conclusions or recommendations expressed in this material are those of the author(s) and do not necessarily reflect the views of the National Science Foundation, or any sponsor.
% We thank Jie Tan and Wenhao Yu for the discussions throughout the project.
% % This work was supported by Korea Institute for Advancement Technology (KIAT) grant funded by 
% This research has been funded by the Industrial Technology Innovation Program (P0028404, development of a product-level humanoid mobile robot for medical assistance equipped with bidirectional customizable human-robot interaction, autonomous semantic navigation, and dual-arm complex manipulation capabilities using large-scale artificial intelligence models) of the Ministry of Industry, Trade and Energy of Korea.
% JK was supported by the National Science Foundation Graduate Research Fellowship under Grant No. DGE-2039655. Any opinion, findings, and conclusions or recommendations expressed in this material are those of the author(s) and do not necessarily reflect the views of the National Science Foundation, or any
% sponsor.
\bibliographystyle{IEEEtran}
\balance
\bibliography{bib}

\clearpage

\section*{APPENDIX}

\vspace{1em}
\subsection{Training detail for RL policy}
% % (**RL training detail**)
The RL policy was trained in simulation using NVIDIA Isaac Gym~\cite{makoviychuk2021isaac}. The policy was learned using PPO~\cite{schulman2017proximalpolicyoptimizationalgorithms} and was trained over approximately three hours using a single NVIDIA RTX 3060 GPU. The policy took onboard robot states and the commanded velocity as inputs and output the leg angles required to reach the commanded velocity. These leg angle targets were tracked using a PD controller. Rewards were collected over 4096 parallel environments where the agent was commanded to achieve forward, sideways, and angular velocities. % (Ranges?)
Each episode terminated either when the robot fell down and contacted the ground or after 10 seconds of running. The general structure for the rewards follows what was used in Legged Gym~\cite{rudin2022learning}. We included an additional reward term $r_{snd}$ to improve the quality of the gait and to attempt to reduce the walking sound:
$$r_{snd}(s_t, a_t)=\sum_i{|f_i|}.$$
% (Describe how?)
% A full list of reward terms and their weights are in the Appendix. (Do we want this?)

\vspace{1em}
\subsection{Robot and human diagram in Part 2. Navigation Movement}
\begin{figure}[h]
    \centering
    \vspace{2em}
    \includegraphics[width=.9\linewidth]{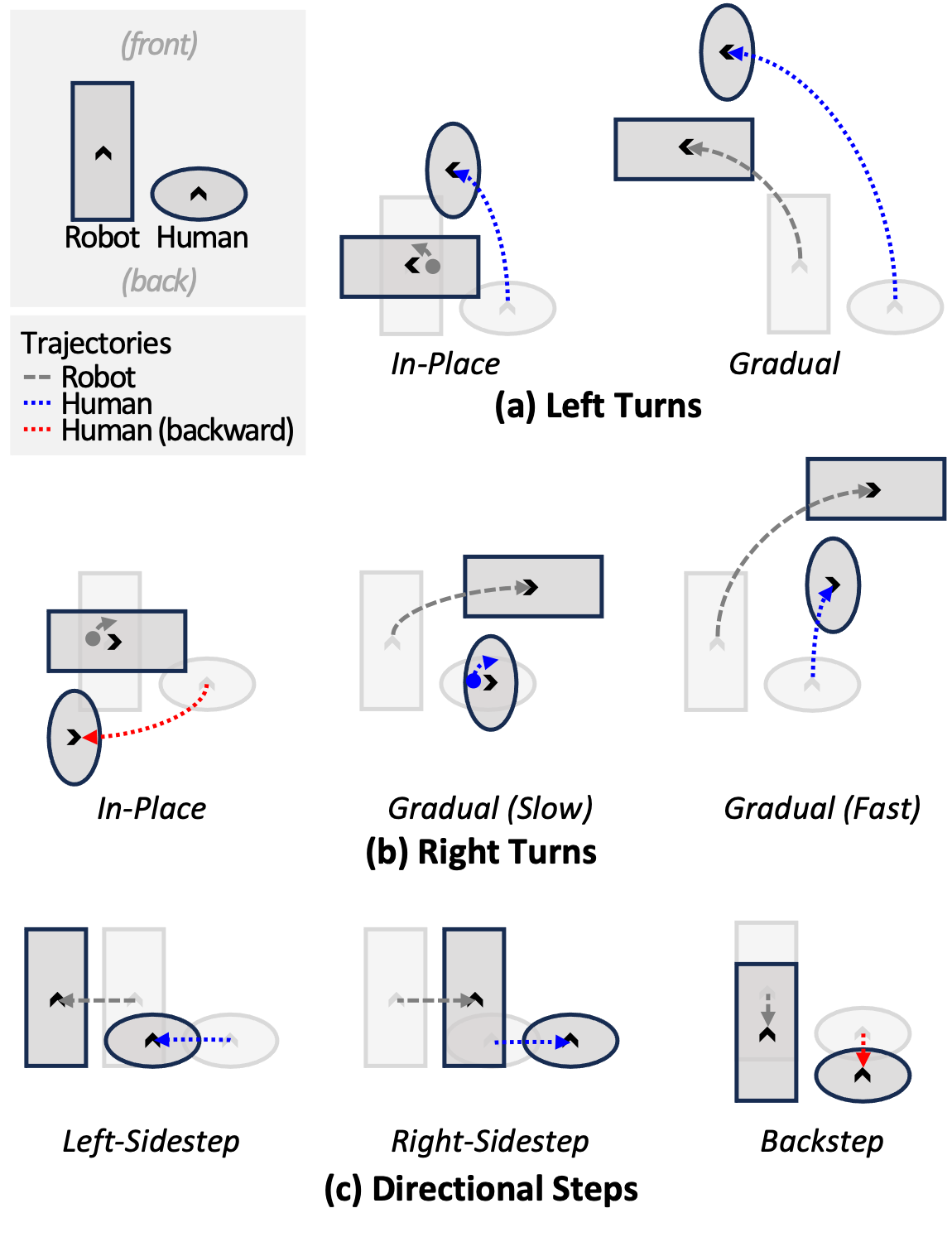}
    \vspace{-1em}
    \caption{Robot and human movement diagram for Part 2.}
    \label{fig:movement-diagram}
\end{figure}

Fig.~\ref{fig:movement-diagram} illustrates the 11 distinct movements the robot performed, along with the corresponding human movements in response.
In general, participants maintained a consistent position relative to the robot.

During left turns, in-place turns required the human to pause, while gradual turns involved the human following along.
The pace of the turns, slow and fast, determined the human's waiting time and the size of the turning circle.

For right turns, in-place turns caused the human to step backward, with the robot being the center of the turning circle. 
Gradual turns showed distinct human movements: gradual-slow turns made the human rotate in place, while gradual-fast turns allowed the human to maintain some forward velocity.

For directional steps, humans followed smoothly for the left and back steps but showed delays for the right steps.

\vspace{1em}
\subsection{Description of items in quantitative questionnaires}
\begin{table*}[h]
\centering
\caption{Descriptions of clarity, security, and frustration for ratings in Part 2 and 3}
\vspace{-.5em}
\begin{tabular}{@{}cc@{}}
\toprule
\textbf{Items} & \textbf{Description}                                                                                                   \\ \midrule
Clarity        & Feeling clear and at ease about the robot’s intended guide, without experiencing confusion, hesitation, or uncertainty \\
Security       & Feeling safe rather than feeling vulnerable or insecure                                                                \\
Frustration    & Feeling disappointed, discouraged, irritated, stressed, or annoyed                                                     \\ \bottomrule
\end{tabular}
\label{tab:q23-three}
\end{table*}
Participants rated three items in Table~\ref{tab:q23-three} across Part 2 (Navigation Movements, Sec.~\ref{sec:exp2-human-movement}) and Part 3 (Bidirectional Communication, Sec.~\ref{sec:p3-bi-comm}) to assess their experiences with different robot behaviors and different types of handles.
The security and frustration items are adopted and modified from the PIADS~\cite{jutai2002psychosocial} and NASA-TLX~\cite{hart2006nasa}, respectively.

% \vspace{3em}
% \subsection{Description of the scale used in Part 4. Explainability}
\begin{table*}[h]
% \begin{table*}[bp]
\centering
% \scriptsize
    \setlength\tabcolsep{1pt}
    \caption{Modified PIADS items and descriptions to compare robotic guide dog experience to user's current mobility aid
}
\vspace{-.5em}
\begin{tabular}{@{}ccc@{}}
\toprule
\textbf{} & \textbf{Item} & \textbf{Description} \\ \midrule
i01 & Ability to adapt to the activities of daily living & Ability to cope with change; ability to make basic tasks more manageable \\
i02 & Ability to participate & Ability to join in activities with other people \\
i03 & Ability to take advantage of opportunities & Ability to act quickly and confidently when there is a chance to improve something in your life \\
i04 & Eagerness to try new things & Feeling adventuresome and open to new experiences \\
i05 & Happiness & Gladness, pleasure; satisfaction with life \\
i06 & Independence & Not dependent on, or not always needing help from, someone or something \\
i07 & Productivity & Able to get more things done in a day \\
i08 & Quality of life & How good your life is \\
i09 & Trust & Feeling trust in mobility aids \\
i10 & Security & Feeling safe rather than feeling vulnerable or insecure \\
i11 & Embarrassment & Feeling awkward or ashamed \\ \bottomrule
\end{tabular}
\label{tab:q4-1-piads}
\end{table*}
Table~\ref{tab:q4-1-piads} outlines the items and descriptions used in the interviews in Part 4 (Explainability, Sec.~\ref{sec:p4-explainability}).
We modified the Psychosocial Impact of Assistive Devices
Scale (PIADS)~\cite{jutai2002psychosocial} based on the short version of the PIADS and the version used for guide robots~\cite{zhang2023follower}, focusing on aspects those are potentially affected by the robotic guide dog with or without audio feature.

% \clearpage
\vspace{1em}
\subsection{Navigation routes}

\begin{figure*}[bp]
    \centering
    \includegraphics[width=.9\linewidth]{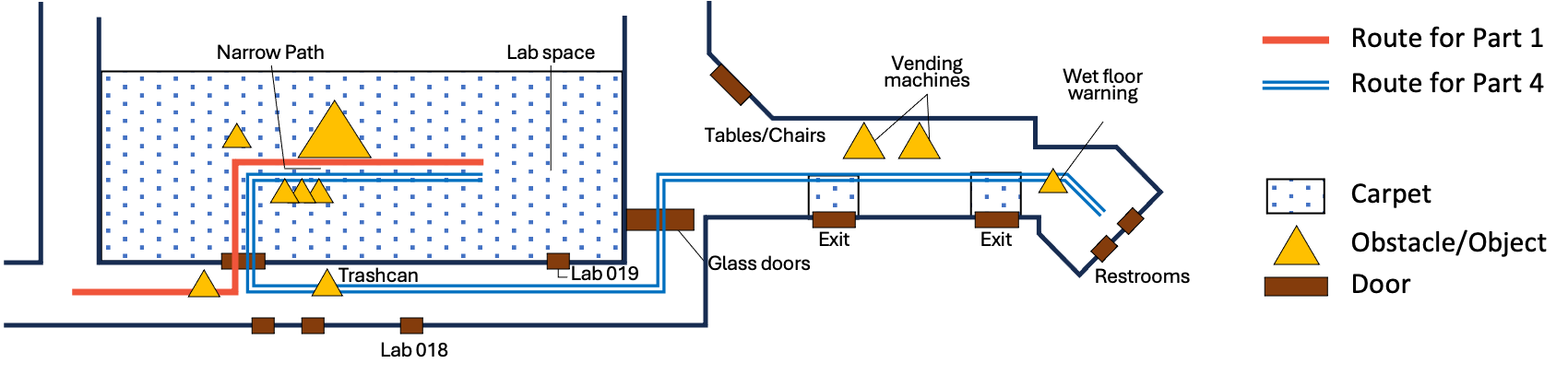}
    \vspace{-1em}
    \caption{Navigation routes for Part 1 and Part 4.}
    \label{fig:map}
\end{figure*}

Fig.~\ref{fig:map} shows the navigation routes for Part 1 (Robot Locomotion, Sec.~\ref{sec:p1-locomotion}) and Part 4 (Explainability, Sec.~\ref{sec:p4-explainability}).
Various obstacles, objects, and doors are marked along the routes, highlighting elements explicitly included in the explainability audio for Part 4.
Route transitions between carpeted and tiled areas are noted to account for environmental changes.

% \section*{ACKNOWLEDGMENT}

% The preferred spelling of the word ÒacknowledgmentÓ in America is without an ÒeÓ after the ÒgÓ. Avoid the stilted expression, ÒOne of us (R. B. G.) thanks . . .Ó  Instead, try ÒR. B. G. thanksÓ. Put sponsor acknowledgments in the unnumbered footnote on the first page.

% \addtolength{\textheight}{-30em}   % --- makes the 

%%%%%%%%%%%%%%%%%%%%%%%%%%%%%%%%%%%%%%%%%%%%%%%%%%%%%%%%%%%%%%%%%%%%%%%%%%%%%%%%


%%%%%%%%%%%%%%%%%%%%%%%%%%%%%%%%%%%%%%%%%%%%%%%%%%%%%%%%%%%%%%%%%%%%%%%%%%%%%%%%

\end{document}